\documentclass[conference]{IEEEtran}
\IEEEoverridecommandlockouts
\usepackage{cite}
\usepackage{amsmath,amssymb,amsfonts}
\usepackage{algorithmic}
\usepackage{graphicx}
\usepackage{textcomp}
\usepackage{xcolor}
\usepackage{amsmath}
\usepackage[bookmarks=false]{hyperref} 
\usepackage{colortbl}
\usepackage{booktabs}
\usepackage[per-mode=symbol,range-phrase=--,range-units=single]{siunitx}
\usepackage{nicefrac}
\usepackage{multirow}
\usepackage{calc}
\usepackage{printlen}
\usepackage[normalem]{ulem}
\usepackage{ragged2e}
\usepackage{amsmath}
\usepackage{graphicx}
\usepackage{amssymb}
\usepackage{multirow}
\usepackage{makecell}
\usepackage{mathrsfs}
\usepackage{threeparttable}
\usepackage{color}
\usepackage{booktabs}
\usepackage{algorithm}
\usepackage{algorithmic}
\def\BibTeX{{\rm B\kern-.05em{\sc i\kern-.025em b}\kern-.08em
   T\kern-.1667em\lower.7ex\hbox{E}\kern-.125emX}}

\usepackage{booktabs}

\usepackage{bbm}

\begin{document}

\title{$\text{S}^2$RC-GCN: A Spatial-Spectral Reliable Contrastive Graph Convolutional Network for Complex Land Cover Classification Using Hyperspectral Images}

\author{
	\IEEEauthorblockN{1\textsuperscript{st} Renxiang Guan}
	\IEEEauthorblockA{\textit{School of Computer} \\
		\textit{China University of Geosciences}\\
		Wuhan, China \\
		grx1126@cug.edu.cn}
\\
	\IEEEauthorblockN{2\textsuperscript{st} Zihao Li}
	\IEEEauthorblockA{\textit{School of Computer} \\
		\textit{China University of Geosciences}\\
		Wuhan, China}
\and
	\IEEEauthorblockN{3\textsuperscript{st} Chujia Song}
	\IEEEauthorblockA{\textit{School of Geography and Information Engineering} \\
		\textit{China University of Geosciences}\\
		Wuhan, China}
\\
\\
	\IEEEauthorblockN{4\textsuperscript{st} Guo Yu}
	\IEEEauthorblockA{\textit{School of Management} \\
		\textit{Wuhan University of Science and Technology}\\
		Wuhan, China}
\and
	\IEEEauthorblockN{5\textsuperscript{st} Xianju Li*}
	\IEEEauthorblockA{\textit{School of Computer} \\
		\textit{China University of Geosciences}\\
		Wuhan, China \\
		ddwhlxj@cug.edu.cn}
\\
	\IEEEauthorblockN{6\textsuperscript{st} Ruyi Feng}
	\IEEEauthorblockA{\textit{School of Computer} \\
		\textit{China University of Geosciences}\\
		Wuhan, China}
\thanks{*Corresponding author}
}

\maketitle

\begin{abstract}
\looseness=-1 Spatial correlations between different ground objects are an important feature of mining land cover research. Graph Convolutional Networks (GCNs) can effectively capture such spatial feature representations and have demonstrated promising results in performing hyperspectral imagery (HSI) classification tasks of complex land. However, the existing GCN-based HSI classification methods are prone to interference from redundant information when extracting complex features. To classify complex scenes more effectively, this study proposes a novel spatial-spectral reliable contrastive graph convolutional classification framework named $\text{S}^2$RC-GCN. Specifically, we fused the spectral and spatial features extracted by the 1D- and 2D-encoder, and the 2D-encoder includes an attention model to automatically extract important information. We then leveraged the fused high-level features to construct graphs and fed the resulting graphs into the GCNs to determine more effective graph representations. Furthermore, a novel reliable contrastive graph convolution was proposed for reliable contrastive learning to learn and fuse robust features. Finally, to test the performance of the model on complex object classification, we used imagery taken by Gaofen-5 in the Jiang Xia area to construct complex land cover datasets. The test results show that compared with other models, our model achieved the best results and effectively improved the classification performance of complex remote sensing imagery.
\end{abstract}


%
%

\section{Introduction}
Land cover information is important for studying biodiversity, human-land relationships, and other issues, as it is a crucial component in resource utilization and environmental protection \cite{liu2017classifying}, \cite{cheng2022IJCNN}. The rapid development of imaging spectroscopy technology probabilizes fine-grained land cover classification (FLCC), and many algorithms work to complete the mapping from remote sensing imagery to pixel-wise labels of land cover \cite{gong2013IJRS}, \cite{whu1}. Specifically, conducting FLCC in complex geographical scenarios is advantageous because of the rich spectral information \cite{sony1}, \cite{sony2}, \cite{sony3}, \cite{sony4}. However, with the gradual improvement in the spectral resolution of hyperspectral image (HSI), the intracategory differences of the same type of ground objects and the redundancy of spectral features have represented severe challenges to FLCC \cite{2022JAGAN}, \cite{chenzhonghao1}, \cite{chenzhonghao2}, \cite{chenzhonghao3}, \cite{chenzhonghao4}.

Deep learning (DL)-based algorithms can extract robust deep features layer by layer and are gaining considerable attention in the field of HSI processing \cite{whu2}, \cite{whu3}, \cite{guan2023pixel}, \cite{liang2022reasoning}.
As typical DL models, convolutional neural networks (CNNs) are widely used to solve HSI classification problems \cite{liu2022CNNIJCNN}, ranging from 1D-CNN to 3D-CNN \cite{38hu1DCNN}, \cite{20chen2016deep}, \cite{20173DCNN}, \cite{hutao1}, \cite{hutao2}, \cite{hutao3}. Hu et al. \cite{38hu1DCNN} first proposed a 1D-CNN to extract the pixel spectral features. To capture spatial features, Chen et al. \cite{20chen2016deep} designed a 2D-CNN model that employed several convolution and pooling layers to obtain nonlinear deep features and used them for HSI classification. Rasti et al. \cite{23rasti2017fusion} proposed a three-stream CNN for LCC using HSI and LIDAR data. In addition, residual networks (ResNets) \cite{24zhong2017spectral}, dense networks \cite{25wang2018fast} and Capsule network \cite{26guan2022classification} have been introduced into HSI tasks and have demonstrated strong performance. However, these methods only focus on extracting image region features with a fixed kernel size, thereby ignoring the long-range spatial relationships of different local regions \cite{chenzhihao1}, \cite{chenzhihao2}, \cite{chenzhihao3}, \cite{chenzhihao4}, \cite{chenzhihao5}. 

To address the disadvantages of CNN, graph convolutional networks (GCNs) have been applied to HSI classification tasks, which can convolve arbitrarily structured graphs and preserve class boundary information \cite{liu2023adaptive}, \cite{liu2023deep}, \cite{wang2022hyperspectral}. Ma et al. \cite{27ma2016graph} first combined spectral and spatial regularization to propose a graph-based semi-supervised spatial-spectral joint feature HSI classification method. Qin et al. \cite{28qin2018spectral} proposed a new framework for semi-supervised learning based on a GCN that considers both the spatial and spectral features of HSI in the filtering operation of the graph.  However, semi-supervised tasks must consider the spatial relationship between all pixels, which incurs a high computational cost. In response to this problem, Wan et al. \cite{29wan2019multiscale} used the SLIC \cite{30achanta2012slic} algorithm to segment an image into superpixels and used them as the nodes of the graph while dynamically updating the graph structure using the graph convolution process. Liu et al. \cite{31liu2020cnn} used a GCN to learn superpixel features, a CNN to learn pixel-level features, and ultimately fused the features. However, these methods use full-batch network learning, which incurs huge memory overhead. In response, Hong et al. \cite{Funet} proposed a mini-batch GCN to integrate the features of a GCN and a CNN in an end-to-end trainable network. Subsequently, an increasing number of methods based on GCN have been introduced into HSI classification research, such as the hypergraph \cite{2ma2021hyperspectral} and spatial-pooling GCN \cite{3zhang2021spectral}.

To enhance the robustness of feature learning, contrastive learning has been incorporated into HSI processing tasks. Huang et al. \cite{huang2022} introduced an innovative approach called 3D swin transformer-based hierarchical contrast learning, which effectively leveraged the multi-scale semantic representation of an image. Lu et al. \cite{lu2022} proposed a novel end-to-end supervised contrastive learning network for spectral–spatial classification. Guan et al. \cite{guan2023contrastive} proposed a contrastive subspace clustering method based on GCN. Yu et al. \cite{yu2023} devised a semi-supervised contrastive loss to exploit the supervision contained in the spectral signatures of image regions \cite{li2023mixture}, \cite{xia3}, \cite{xia5}. 

Nevertheless, current GCN-based contrastive methods still contain some disadvantages. On the one hand, the positive samples generated by these methods are typically limited to one, whereas negative samples are randomly chosen, introducing a certain level of randomness and the potential for misclassification. This randomness diminishes the model's capability to extract features to some extent \cite{shen1}, \cite{shen2}, \cite{xia1}. Therefore, it is imperative to develop a reliable contrastive method during feature propagation between adjacent layers of a GCN. On the other hand, an adjacency matrix is known to represent the topology of the graph data \cite{xia2}, \cite{xia4}. However, most GCN-based methods use only the shallow spatial or spectral information of HSI to construct graphs, which can only represent pairwise relationships and cannot discover the underlying complex structures in the complete land cover imagery \cite{shen3}, \cite{shen4}, \cite{shen5}.

To alleviate these deficiencies, we propose a novel spatial- spectral reliable contrastive graph convolutional network ($\text{S}^2$RC-GCN) for complex land cover classification. In the model, 1D-encoder and 2D-encoder learn high-level spectral and spatial features respectively from HSI to further construct deep graphs, and then feed the generated graphs into GCN for further classification tasks. In particular, we propose a reliable contrastive GCN, which uses nodes with true labels or predictive labels with high thresholds for contrastive learning to further learn more efficient graph representations. To test the effectiveness of the model in complex ground object classification tasks, this paper uses two complex land cover datasets captured by the Gaofen-5 satellite (GF-5) and a general dataset. The main contributions of this paper are as follows:
\begin{enumerate}
    \item This paper designs a new CNN-GCN framework named $\text{S}^2$RC-GCN. It fuses spatial and spectral features by 1D-encoder and 2D-encoder, and 2D-encoder includes an attention model to automatically extract important information which can effectively capture deep features for robust graph construction.
    \item The supervised contrastive loss and the enhancement of reliable samples are crafted to leverage the supervision signals inherent in the HSI, avoiding the noise introduced by the data augmentation process.
    \item To better test the robustness and generalizability of the model, the paper uses two complex land cover datasets captured by the GF-5 and a general HSI dataset.
\end{enumerate}

%
%

\section{Method}
\label{ch3:motivation}

A flowchart of the proposed $\text{S}^2$RC-GCN system is shown in the Figure \ref{fig:main}. In this section, we introduce the details of each part and illustrate the implementation. We first introduce the basic feature extraction model included in the algorithm; then the patch-level spatial information and pixel-level spectral information are extracted and fused; the third section introduces how to use the fused features and spectral features to construct robust graph adjacency matrices; Afterward, the fourth section details the construction of reliable contrastive graph convolution.

\begin{figure*} \label{fig:main}
\begin{center}
  \includegraphics[width=\linewidth]{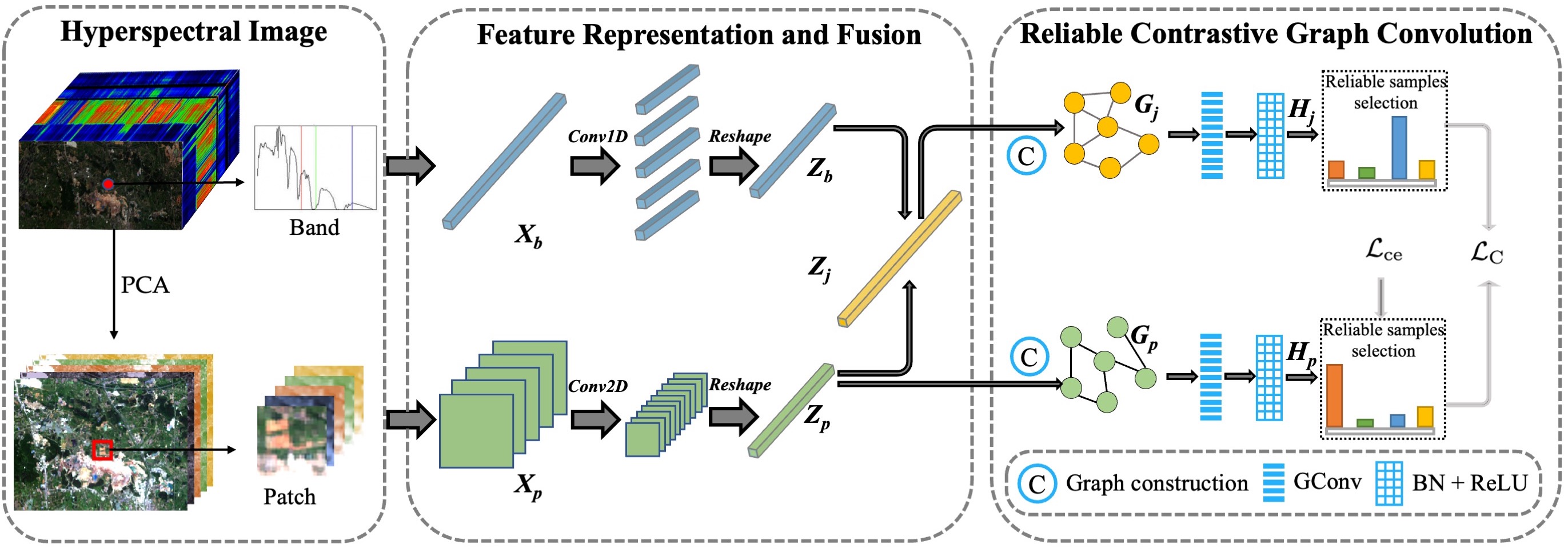}
\end{center}
\caption{Architecture of the proposed $\text{S}^2$RC-GCN. The model comprises three main components. Initially, it acquires the spatial and spectral features of the HSI, followed by utilizing 1D-CNN and 2D-CNN for feature extraction and fusion. Lastly, spatial features and the fused representation are employed for composition and reliable contrastive learning.}
\label{fig:main}
\end{figure*}

\subsection{Basic Model}
\label{se:A}
\textbf{2D-CNN (SE-ResNet):} Because the CNN uses a convolution kernel to fuse the spatial information on the local receptive field and the information on the channel dimension to construct the feature map, it causes multiple problems simultaneously \cite{LIrui}, \cite{applied}. Different feature channels are further used with equally important weights to create global irrelevance, and the features are propagated through the network, thus affecting their accuracy. Therefore, the feature extraction network designed in this study is SE-ResNet. Before Squeeze-and-Excitation model is applied, we use the principal component analysis (PCA) \cite{2002PCA} to reduce the band dimension of HSI to reduce redundant spectral information. Figure \ref{fig:2} shows a schematic of the SE-ResNet network structure. SE-Net is implemented in two steps: 1) global information embedding and 2) adaptive rescaling. In global information embedding, the output of the CNN compresses the features along the spatial dimension through the global pooling layer. In adaptive rescaling, a gate with a sigmoid activation function is selected to obtain a normalized weight between zero and one, which is the scale weight factor used to characterize the importance of each channel. Finally, the normalized weights were assigned to the features of each channel through a scale operation to complete the recalibration of the original features in the channel dimension.

\textbf{GCN:} HSI can be represented as a graph $G=(V,E,X)$, where $V=\{v_i\}_{i=1,2,\ldots}$ represents the set of pixels, $N$ denotes the number of nodes, $E$ is the set of concatenated edges, and $X \in \mathbb{R}^{N \times F}$ is the feature matrix containing attribute vectors for each node. $A \in \mathbb{R}^{N \times N}$ represents the corresponding adjacency matrix of the network, where $A_{ij} = 1$ if there is a concatenated edge between nodes $v_i$ and $v_j$, otherwise $A_{ij} = 0$. $D \in \text{diag}(d_1, d_2, d_3, \ldots, d_i)$ represents the corresponding degree matrix of the graph $G$, where $d_i = \sum_{j=1}^{N} A_{ij}$.

GCN extend traditional CNN to the graph domain, currently categorized into two types: null-domain GCNs and frequency-domain GCNs. This paper predominantly adopts frequency-domain GCNs, specifically the conventional type of GCN. We utilize frequency-domain GCNs to define the convolution operation by decomposing the graph signal in the frequency domain and then applying a spectral filter to the frequency domain components. The layer-by-layer propagation rule for graph convolution can be expressed as follows:
\begin{equation}H^{(l+1)}=\sigma(\widetilde{D}^{-\frac12}\widetilde{A}\widetilde{D}^{-\frac12}H^{(l)}W^{(l)})\end{equation}
where $\tilde{A} = A + I_N$ represents the adjacency matrix of the added self-looped undirected graph $G$ derived from the original adjacency matrix, and $I_N$ is the unit matrix. The output matrix of layer $l$ is denoted as $h^{(l)}$, with $H^{(0)}=X$ and $l={0,1,\ldots,L-1}$ indicating the number of convolutional layers. $w^{(l)}$ refers to the trainable weight matrix for a specific layer. The activation function $\sigma(\cdot)$ exemplifies ReLU in the experiments.

\subsection{Joint Spatial-Spectral Feature Extraction}

Adjacent pixels are often associated with the same class in the HSI. Existing methods typically employ a 2D-CNN to extract features from patches. However, this approach may lead to a reduction in the contribution of spectral information to the central region of the image, resulting in reduced robustness against noisy samples. To address this issue, we introduced a 1D-CNN into the network to fully leverage the spectral information of the HSI. The 1D-CNN extracts spectral features from bands, whereas the 2D-CNN extracts spatial-spectral features from patches. These two sets of features are concatenated to create a more resilient joint spatial-spectral features representation. To suppress the influence of useless information, we designed a SE-ResNet network that forced the model to focus on the feature information of useful channels, the details of which can be found in Section \ref{se:A}.

$X_b$ and $X_p$ represent the spectral and spatial data of the HSI, respectively, where $X_b \in \mathbb{R}^{d \times N}$, $d$ denotes the dimension of the samples. Additionally, a square with side length $w$ is used to capture the spatial features around each pixel. The spectral dimension is reduced to $p$ using PCA to obtain the initial spatial data $X_p \in \mathbb{R}^{\hat{d} \times w \times w \times N}$. Next, we process $X_b$ with the 1D-CNN to obtain the spectral feature $Z_b\in\mathbb{R}^{l_b \times N}$, where $l_b$ represents the length of features. Simultaneously, we use the 2D-CNN to obtain the spatial feature $Z_p$. The encoding operations are defined as follows:
\begin{equation}
\begin{cases}
Z_b=f_b(X_b)\in\mathbb{R}^{l_b \times N}\\
Z_p=f_p(X_p)\in\mathbb{R}^{l_p \times w_l \times w_l \times N},
\end{cases}
\end{equation}
where $f_{b}\left(\cdot\right)$ and $f_{p}\left(\cdot\right)$ denote the mapping operations of the 1D-CNN and 2D-CNN, respectively. Because feature $Z_p$  is a three-dimensional array, we flattened it to obtain a feature vector of length $l_p$ as well as a new feature representation, $Z_p \in \mathbb {R}^{n \times l_p}$. Finally, we concatenate $Z_p$ and $Z_b$ to obtain the joint spatial-spectral feature $Z_j \in \mathbb{R}^{n \times l}$, where $l=l_b + l_p$.

\subsection{Graph Construction}

Traditionally, the use of shallow features of HSI to construct graphs is not robust and cannot be updated through the learning process. In this study, we used the spatial feature $Z_p$ and spatial-spectral joint feature $Z_j$ to construct the nodes of the graphs. Edges are defined as potential interactions between two nodes, with the assumption that nodes with smaller distances are more likely to interact. As a node represents a spatial region in the original HSI, the constructed graph represents the spatial correlation between different regions of the image. The Euclidean distance defined in Eq. (\ref{eq:3}) is used. First, the correlation $dis\left(z^i,z^j\right)$ between all nodes is calculated to obtain the distance matrix $Dis\in \mathbb{R}^{N\times N}$.

\begin{equation} \label{eq:3}
Dis\left(Z\right)=\sqrt{\Sigma_{i=1}^{N}\Sigma_{j=1}^{N}\left(z^i-z^j\right)^{2}}.
\end{equation}

Then, the K nodes closest to each node are selected as neighbors, and edges are generated between these nodes to obtain the structure of the graph. The adjacency matrix $A \in \mathbb{R}^{N \times N}$ is defined as:
\begin{equation}
A_{ij}=\begin{cases}1,&z^i\in \mathcal{N}_k\left(z_i\right)~or~z^j\in \mathcal{N}_k\left(z_i\right)~\\0,&Otherwise\end{cases}.
\end{equation}
where $\mathcal{N}_k\left(z_i\right)$ represents the set of neighbors. Through the above process, we input spatial features $Z_p$ and joint features $Z_j$ to obtain their adjacency matrices $A_p$ and $A_j$, respectively. Combining the features and their corresponding adjacency matrices we obtain two graphs, $G_p=(Z_p, A_p)$ and $G_j=(Z_j, A_j)$. These two graphs then flow into two branches. and are processed using a graph convolution module. The mathematical description of this step is as follows:
\begin{equation}
\begin{cases}
G_p^{\prime}=\text{GConv}(G_p)+w_p\\G_j^{\prime}=\text{GConv}(G_j)+w_j&
\end{cases}
\end{equation}
where $\text{GConv}(\cdot)$ and $w$ represent the graph convolution operation and corresponding bias, respectively. The resulting $G_p^{\prime}$ and$G_j^{\prime}$ are then fed into the BN+ReLU module, generating nonlinear graph features $G_p^{\prime\prime}$ and $G_j^{\prime\prime}$ as follows:

\begin{equation}
\left.\left\{\begin{aligned}G_p^{\prime\prime}&=ReLU\big(BN(G_p^{\prime})\big)\\G_j^{\prime\prime}&=ReLU\big(BN(G_j^{\prime})\big)\end{aligned}\right.\right.
\end{equation}

Finally, we obtain graph features notated $H_j$ and $H_p$ that aggregate neighbor information.

\begin{figure} \label{fig:2}
\begin{center}
  \includegraphics[width=\linewidth]{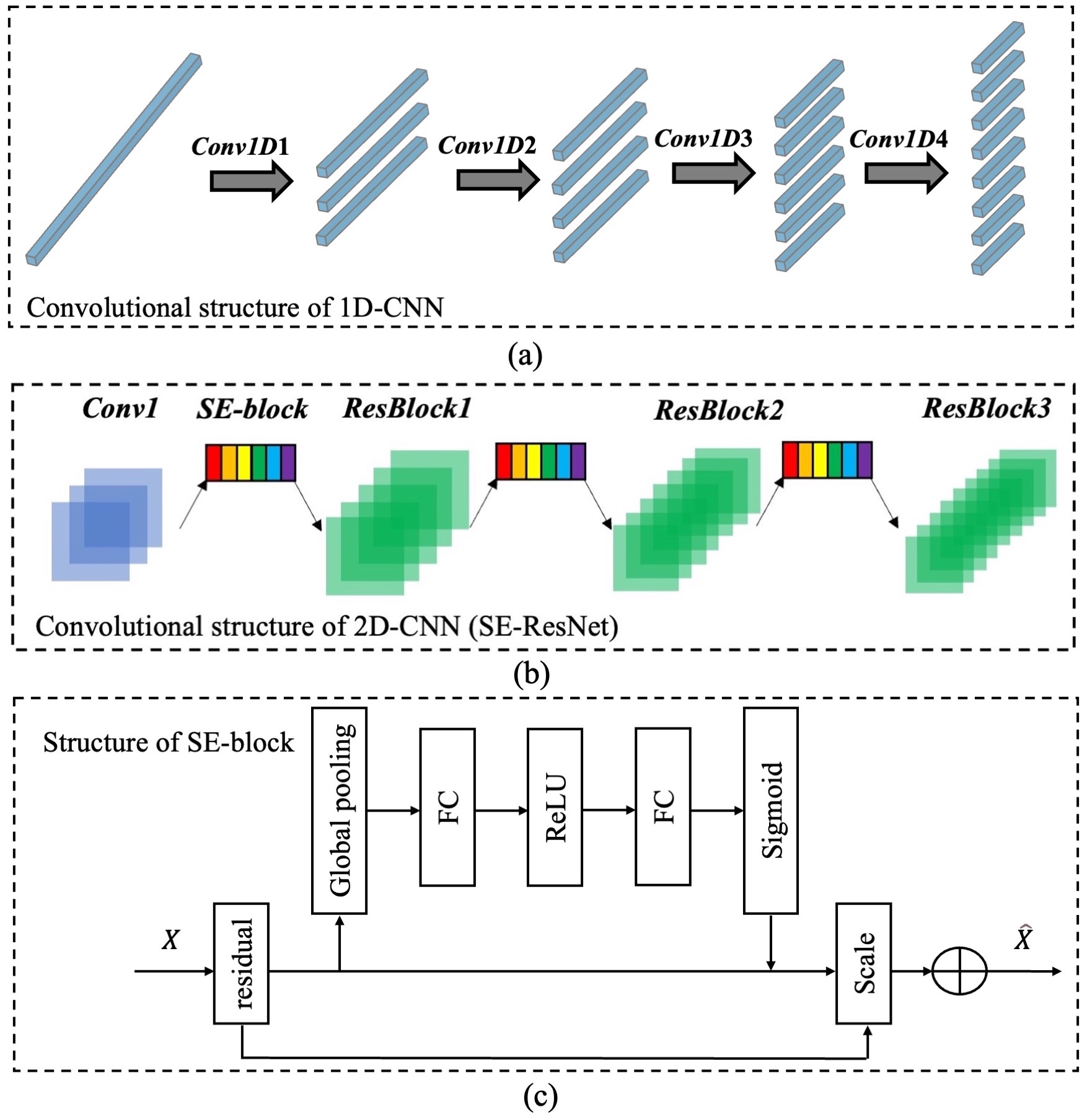}
\end{center}
\caption{Structural diagram of 1D-CNN and 2D-CNN, which also includes a detailed construction of the Squeeze-and-Excitation module. (a) 1D-CNN; (b) 2D-CNN; (c) SE-block}
\label{fig:2}
\end{figure}

\subsection{Reliable Contrastive GCN}

Supervised contrastive loss provides rich supervisory information for contrastive learning using only unlabeled samples, assisting the model in learning better node representations and further improving model performance. During the learning process, an evaluation function is employed to measure the similarity of node encoding features. The goal is to maximize the similarity between nodes of the same category while minimizing the similarity between nodes of different categories, guiding the learning process.

Existing contrastive learning methods often use data augmentation to construct positive and negative sample pairs, but the augmentation process is prone to introducing noise. Therefore, this paper utilizes spatial feature $H_p$ and spatial-spectral feature $H_j$ for contrastive learning. Both features are extracted from hyperspectral data, avoiding the noise introduced by the data augmentation process. Assuming $x^i$ is a labeled node, so $h_j^i$ and $h_p^i$ are the corresponding spatial and joint features, respectively. The positive example is a labeled node in the spatial feature with the same class, and the negative example is a labeled node with a different class. 

\begin{table*}[!h]
\caption{Classification Results (OA/NMI/KAPPA) of Various Methods on \textbf{Jiang Xia} Datasets. The Best Results are Shown in Bold.}
\label{tb_jiangxia}
\scalebox{0.92}{
\renewcommand\arraystretch{1.2}
\begin{tabular}{ccccccccccc}
\hline
Class No. & 1D-CNN & 2D-CNN & 3D-CNN & ResCapsNet & FuNet & SAGE-A & MDGCN & $\text{F}^2\text{HNN}$ & Ours \\ \hline
1 & 59.47 $\pm$ 2.11 & 63.16 $\pm$ 0.52 & 75.51 $\pm$ 0.72 & 76.35 $\pm$ 0.56 & 80.59 $\pm$ 0.14 & 80.96 $\pm$ 1.12 & \textbf{89.58 $\pm$ 1.32} & 90.18 $\pm$ 0.72 & 88.27 $\pm$ 1.28 \\
2 & 77.35 $\pm$ 1.28 & 82.44 $\pm$ 0.89 & 85.29 $\pm$ 1.41 & 91.42 $\pm$ 0.54 & 88.07 $\pm$ 0.42 & 73.16 $\pm$ 0.74 & 75.57 $\pm$ 0.51 & 90.70 $\pm$ 0.58 & \textbf{97.04 $\pm$ 0.79} \\
3 & 95.87 $\pm$ 1.01 & 98.92 $\pm$ 0.23 & 94.23 $\pm$ 1.08 & \textbf{100.0 $\pm$ 0.00} & 99.35 $\pm$ 0.11 & 75.32 $\pm$ 0.87 & 84.51 $\pm$ 2.75 & 95.19 $\pm$ 0.21 & 93.42 $\pm$ 0.85 \\
4 & 94.61 $\pm$ 1.89 & 95.96 $\pm$ 1.41 & 91.03 $\pm$ 0.69 & 95.76 $\pm$ 0.27 & \textbf{96.88 $\pm$ 0.57} & 82.88 $\pm$ 0.74 & 93.54 $\pm$ 2.68 & 86.25 $\pm$ 0.46 & 92.26 $\pm$ 0.72 \\
5 & 71.02 $\pm$ 0.22 & 61.61 $\pm$ 0.57 & 72.09 $\pm$ 1.77 & 55.92 $\pm$ 0.80 & 46.94 $\pm$ 0.78 & 45.52 $\pm$ 0.28 & 71.48 $\pm$ 0.45 & 50.73 $\pm$ 0.55 & \textbf{75.24 $\pm$ 1.20} \\
6 & 80.66 $\pm$ 0.17 & 74.22 $\pm$ 0.83 & 76.72 $\pm$ 0.94 & 85.24 $\pm$ 1.26 & 68.93 $\pm$ 0.87 & 66.54 $\pm$ 0.81 & 73.83 $\pm$ 1.66 & 70.48 $\pm$ 0.26 & \textbf{88.21 $\pm$ 0.81} \\
7 & 73.59 $\pm$ 1.33 & 71.56 $\pm$ 0.45 & 75.20 $\pm$ 0.47 & 76.44 $\pm$ 0.75 & 76.23 $\pm$ 0.92 & 54.41 $\pm$ 0.43 & 81.50 $\pm$ 3.05 & 72.89 $\pm$ 0.88 & \textbf{82.77 $\pm$ 0.42} \\\hline
OA & 79.59 $\pm$ 1.79 & 80.44 $\pm$ 0.69 & 82.82 $\pm$ 0.53 & 82.80 $\pm$ 0.25 & 79.42 $\pm$ 0.77 & 83.43 $\pm$ 0.98 & 85.28 $\pm$ 1.08 & 79.19 $\pm$ 1.41 & \textbf{87.39 $\pm$ 1.24} \\
AA & 81.68 $\pm$ 0.33 & 78.39 $\pm$ 0.61 & 80.23 $\pm$ 0.68 & 81.35 $\pm$ 1.03 & 78.34 $\pm$ 0.79 & 80.90 $\pm$ 1.48 & 83.03 $\pm$ 1.26 & 78.26 $\pm$ 1.24 & \textbf{86.69 $\pm$ 0.88} \\
Kappa & 80.60 $\pm$ 0.28 & 76.84 $\pm$ 0.96 & 79.64 $\pm$ 0.72 & 79.69 $\pm$ 0.24 & 79.13 $\pm$ 0.41 & 81.71 $\pm$ 0.76 & 82.31 $\pm$ 1.07 & 75.31 $\pm$ 1.49 & \textbf{84.90 $\pm$ 1.74} \\ \hline
\end{tabular}}
\end{table*}

To increase the quantity of positive and negative sample pairs while ensuring the reliability of contrastive learning, this paper employs a one-layer perceptron to output the classification probability before constructing sample pairs. If the probability is greater than a certain threshold, it is included in the labeled set. Therefore, we make full use of scarce but valuable labeled samples, using supervised contrastive loss to provide additional supervisory signals for learning node representations. The supervised contrastive loss is expressed as follows:
\begin{equation} \label{eq:5}
\begin{aligned}
&\ell(h_j^i,h_p^i)~= -\lg \\
&\left[\frac{\sum_{m=1}^{M}1_{[y^{i}=\phi(y^m)]}\exp(h_{j}^{i},h_{p}^{m})}{\sum_{m=1}^{M} (1_{[y^{i}=\phi(y^m)]}\exp(h_{j}^{i},h_{p}^{m})+1_{[y^{i}\neq \phi(y^m)]}\exp(h_{j}^{i},h_{p}^{m}))}\right],
\end{aligned}
\end{equation}
where $\begin{aligned}1_{i \neq m}=\{0,1\}\end{aligned}$ is an indicator function. $\phi(y^m)= \left\{ y^{i}=y^m~\text{or}~ y^{i}=f(h^m)\right\} $, where $f(h^m)$ represents obtaining the predicted label of $h^m$ and $M$ is the number of the set $\phi(y^m)$. There are two critical parameters in this algorithm. One is the threshold for label prediction, which is set to 0.99 in this article to ensure label reliability. The other is the temperature coefficient in contrastive learning, which is consistently set to 1 in all experiments. The overall objective for minimizing is the
average of Eq.~(\ref{eq:5}) for all given positive pairs, which is given by the following expression:
\begin{equation}
\mathcal{L}_{\mathrm{C}}=\frac1{2n}\sum_{i=1}^n[\ell(h^i_j,h_p^i)+\ell(h_p^i,h_j^i)].
\end{equation}

Next, the actual output of the network is measured against the true labels using the cross-entropy loss $\mathcal{L}_{\mathrm{ce}}$, which signifies the difference between the predicted output and the actual labels. Combining the contrastive loss and the cross-entropy loss, the overall output of the network is obtained, and the overall objective loss function is expressed as follows:
\begin{equation}
\mathcal{L}~=~\mathcal{L}_{\mathrm{C}}~+~ \mathcal{L}_{\mathrm{ce}}.
\end{equation}

\begin{table*}[!t]
\caption{Classification Results (OA/NMI/KAPPA) of Various Methods on \textbf{Xin Jiang} Datasets. The Best Results are Shown in Bold.}
\label{tb_xinjiang}
\scalebox{0.92}{
\renewcommand\arraystretch{1.2}
\begin{tabular}{ccccccccccc} \hline
Class & 1D-CNN & 2D-CNN & 3DCNN & ResCapsNet & FuNet & SAGE-A & MDGCN & $\text{F}^2\text{HNN}$ & Ours \\ \hline
1 & 69.42 $\pm$ 1.37 & 69.62 $\pm$ 0.93 & 66.26 $\pm$ 0.37 & 77.66 $\pm$ 0.39 & \textbf{94.08 $\pm$ 0.71} & 91.87 $\pm$ 0.26 & 82.47 $\pm$ 0.93 & 92.19 $\pm$ 1.25 & 88.92 $\pm$ 0.37 \\
2 & 58.83 $\pm$ 0.76 & 63.52 $\pm$ 0.94 & 87.47 $\pm$ 0.75 & 85.29 $\pm$ 0.58 & 87.42 $\pm$ 0.53 & 77.62 $\pm$ 0.38 & 83.50 $\pm$ 0.29 & 86.98 $\pm$ 0.86 & \textbf{88.20 $\pm$ 0.58} \\
3 & 81.65 $\pm$ 0.27 & 90.61 $\pm$ 1.27 & 93.73 $\pm$ 0.89 & \textbf{96.27 $\pm$ 0.62} & 93.04 $\pm$ 1.27 & 93.59 $\pm$ 1.36 & 89.63 $\pm$ 0.87 & 94.12 $\pm$ 1.83 & 93.41 $\pm$ 1.28 \\
4 & \textbf{40.25 $\pm$ 1.89} & 38.82 $\pm$ 0.48 & 30.15 $\pm$ 1.24 & 31.67 $\pm$ 0.74 & 25.49 $\pm$ 0.20 & 28.29 $\pm$ 1.31 & 19.60 $\pm$ 1.78 & 26.79 $\pm$ 0.54 & 27.75 $\pm$ 1.16 \\
5 & 14.55 $\pm$ 0.73 & 16.57 $\pm$ 1.63 & 32.65 $\pm$ 0.31 & 28.85 $\pm$ 0.89 & 26.48 $\pm$ 0.49 & 29.75 $\pm$ 1.08 & 31.38 $\pm$ 0.96 & 28.44 $\pm$ 0.57 & \textbf{36.28 $\pm$ 1.69} \\
6 & 73.26 $\pm$ 0.84 & 81.04 $\pm$ 0.86 & 83.64 $\pm$ 0.58 & \textbf{90.13 $\pm$ 0.41} & 86.38 $\pm$ 0.44 & 75.26 $\pm$ 0.25 & 80.27 $\pm$ 0.44 & 84.08 $\pm$ 0.71 & 85.65 $\pm$ 0.29 \\
7 & 46.31 $\pm$ 1.49 & 49.61 $\pm$ 0.74 & 65.29 $\pm$ 0.38 & 65.61 $\pm$ 0.28 & 71.19 $\pm$ 1.12 & 73.84 $\pm$ 0.42 & 70.36 $\pm$ 0.86 & 72.24 $\pm$ 1.47 & \textbf{76.48 $\pm$ 0.90} \\ \hline
OA & 54.81 $\pm$ 0.74 & 58.50 $\pm$ 0.71 & 67.02 $\pm$ 0.70 & 70.88 $\pm$ 0.70 & 68.80 $\pm$ 0.34 & 65.60 $\pm$ 0.75 & 64.81 $\pm$ 0.89 & 68.47 $\pm$ 1.74 & \textbf{71.42 $\pm$ 0.56} \\
F1-score & 54.80 $\pm$ 0.91 & 58.46 $\pm$ 0.56 & 64.35 $\pm$ 0.21 & 66.27 $\pm$ 0.82 & 63.07 $\pm$ 0.98 & 64.07 $\pm$ 1.42 & 62.27 $\pm$ 0.43 & 64.56 $\pm$ 2.17 & \textbf{68.98 $\pm$ 0.87} \\
Kappa & 47.28 $\pm$ 0.41 & 51.54 $\pm$ 0.29 & 61.32 $\pm$ 0.44 & 62.32 $\pm$ 0.39 & 60.38 $\pm$ 0.44 & 61.38 $\pm$ 1.03 & 58.96 $\pm$ 0.64 & 63.59 $\pm$ 2.03 & \textbf{65.02 $\pm$ 0.76} \\ \hline
\end{tabular}}
\end{table*}

\section{Experiments}
\subsection{Datasets}
In this experiment, to comprehensively assess model performance, the paper utilizes a diverse set of datasets, including the complex land cover dataset, Jiang Xia, Xin Jiang and a generalized hyperspectral datasets, Salinas for evaluation.

\textbf{Jiang Xia dataset:} captured by the GF-5 satellite, comprises 330 bands with a spatial resolution of 30 m \cite{2022JAGAN}. It encompasses seven classes: surface-mined area, road, water, crop land, forest land, and construction land, totaling 121,303 samples. For this study, 200 samples from each class are selected for training, while the remaining samples are reserved for testing. We follow the spatially weakly dependent dataset of the literature \cite{26guan2022classification} for our experiments.

\textbf{Xin Jiang dataset:} is captured by the GF-5 satellite, as is the Jiang Xia dataset. The images are for the lli Kazakh Autonomous Prefecture in Xinjiang. We also use the spatially weakly dependent dataset of Xinjiang from literature \cite{26guan2022classification} for testing.

\textbf{Salinas dataset:} captured by the AVIRIS sensor in 1992, exhibits dimensions of 512 × 217 × 224 after processing, with 512 × 217 pixels and 224 bands. The Salinas dataset includes a total of 16 classes. For this study, 30 samples from each class are selected for training, and the remaining samples are earmarked for testing.

\subsection{Experimental Setup}
\textbf{Compared Methods.} To facilitate a comprehensive comparison of models, this paper selects nine deep learning models for evaluation. These comprise four CNN-based models: 1D-CNN \cite{38hu1DCNN}, 2D-CNN \cite{20chen2016deep}, 3D-CNN \cite{20173DCNN}, and ResCapsNet \cite{26guan2022classification}, and four GNN-based models: FuNet \cite{Funet}, $\text{F}^2$HNN \cite{F2HNN}, graph sample and aggregate attention (SAGE-A) \cite{2022SAGE}, and MDGCN \cite{29wan2019multiscale}. Each of these models is computed with optimized parameters.

\textbf{Evaluation Metrics.} In this paper, we employ three assessment indicators: Overall Accuracy (OA), Average Accuracy (AA), and Kappa. These indicators range from 0 to 1, with larger values indicating superior model performance.

\begin{table*}[!t]
\caption{Classification Results (OA/NMI/KAPPA) of Various Methods on \textbf{Salinas} Datasets. The Best Results are Shown in Bold.}
\label{tb_salinas}
\scalebox{0.92}{
\renewcommand\arraystretch{1.2}
\begin{tabular}{ccccccccccc} \hline
Class No. & 1D-CNN & 2D-CNN & 3D-CNN & ResCapsNet & FuNet & SAGE-A & MDGCN & $\text{F}^2\text{HNN}$ & Ours \\ \hline
1 & 48.81 $\pm$ 0.12 & \textbf{100.0 $\pm$ 0.00} & 88.45 $\pm$ 2.15 & 99.40 $\pm$ 0.42 & 99.34 $\pm$ 0.53 & 99.78 $\pm$ 0.21 & 99.98 $\pm$ 0.03 & 98.19 $\pm$ 1.03 & \textbf{100.0 $\pm$ 0.00} \\
2 & 72.53 $\pm$ 1.13 & 65.97 $\pm$ 4.60 & 87.08 $\pm$ 3.67 & 99.46 $\pm$ 0.39 & 99.17 $\pm$ 0.21 & \textbf{100.0 $\pm$ 0.00} & 99.90 $\pm$ 0.28 & 96.52 $\pm$ 1.19 & \textbf{100.0 $\pm$ 0.00} \\
3 & 67.15 $\pm$ 2.13 & 71.53 $\pm$ 3.19 & 80.03 $\pm$ 2.13 & 98.58 $\pm$ 1.42 & 96.54 $\pm$ 1.39 & 99.45 $\pm$ 0.35 & 99.80 $\pm$ 0.21 & 94.36 $\pm$ 2.13 & \textbf{100.0 $\pm$ 0.00} \\
4 & 51.70 $\pm$ 1.89 & 93.24 $\pm$ 3.19 & 75.31 $\pm$ 5.42 & 99.70 $\pm$ 0.17 & 97.32 $\pm$ 0.27 & \textbf{100.0 $\pm$ 0.00} & 97.49 $\pm$ 2.16 & 99.18 $\pm$ 0.28 & 99.13 $\pm$ 0.46 \\
5 & 86.93 $\pm$ 1.47 & 93.78 $\pm$ 1.01 & 90.15 $\pm$ 3.17 & \textbf{98.90 $\pm$ 1.01} & 98.75 $\pm$ 0.89 & 98.75 $\pm$ 1.17 & 97.96 $\pm$ 0.77 & 99.77 $\pm$ 0.31 & 93.23 $\pm$ 2.59 \\
6 & 90.65 $\pm$ 2.01 & 89.74 $\pm$ 2.32 & 96.34 $\pm$ 2.38 & 99.57 $\pm$ 0.38 & 99.18 $\pm$ 0.27 & 89.72 $\pm$ 3.62 & 99.10 $\pm$ 1.67 & \textbf{100.0 $\pm$ 0.00} & 96.67 $\pm$ 1.37 \\
7 & 60.91 $\pm$ 2.11 & 98.26 $\pm$ 1.64 & 71.79 $\pm$ 4.62 & 99.50 $\pm$ 0.42 & 98.67 $\pm$ 1.30 & \textbf{100.0 $\pm$ 0.00} & 98.18 $\pm$ 1.49 & 99.99 $\pm$ 0.01 & 99.29 $\pm$ 0.34 \\
8 & 94.31 $\pm$ 0.89 & 81.98 $\pm$ 4.32 & 97.78 $\pm$ 1.14 & 75.59 $\pm$ 6.72 & 72.38 $\pm$ 3.98 & 85.25 $\pm$ 5.27 & 92.78 $\pm$ 4.61 & 87.79 $\pm$ 4.89 & \textbf{99.74 $\pm$ 0.13} \\
9 & 40.42 $\pm$ 3.71 & 99.47 $\pm$ 0.51 & 64.31 $\pm$ 6.33 & 99.75 $\pm$ 0.19 & 97.29 $\pm$ 0.61 & 95.31 $\pm$ 2.41 & \textbf{100.0 $\pm$ 0.00} & 99.67 $\pm$ 0.33 & 92.98 $\pm$ 2.41 \\
10 & 71.60 $\pm$ 1.69 & 75.21 $\pm$ 2.75 & 86.74 $\pm$ 3.26 & 94.29 $\pm$ 1.90 & 91.44 $\pm$ 1.64 & 97.18 $\pm$ 0.82 & \textbf{98.31 $\pm$ 1.29} & 96.53 $\pm$ 2.55 & 93.91 $\pm$ 1.19 \\
11 & 77.73 $\pm$ 2.11 & 61.24 $\pm$ 2.68 & 83.02 $\pm$ 2.97 & 97.57 $\pm$ 0.91 & 96.82 $\pm$ 0.78 & 96.36 $\pm$ 1.42 & 99.39 $\pm$ 0.55 & 99.19 $\pm$ 0.21 & \textbf{99.76 $\pm$ 0.17} \\
12 & 60.83 $\pm$ 2.57 & 79.98 $\pm$ 0.45 & 98.10 $\pm$ 0.57 & \textbf{99.99 $\pm$ 0.01} & 99.21 $\pm$ 0.32 & 99.75 $\pm$ 0.12 & 99.01 $\pm$ 0.78 & 93.78 $\pm$ 1.01 & 95.30 $\pm$ 0.99 \\
13 & 95.87 $\pm$ 2.33 & 96.73 $\pm$ 1.66 & 96.83 $\pm$ 1.13 & \textbf{99.95 $\pm$ 0.05} & 97.29 $\pm$ 0.86 & 96.39 $\pm$ 2.28 & 97.59 $\pm$ 1.32 & 97.78 $\pm$ 1.14 & 90.15 $\pm$ 1.26 \\
14 & 92.72 $\pm$ 5.91 & 91.50 $\pm$ 3.05 & 88.05 $\pm$ 3.55 & 98.57 $\pm$ 0.28 & 95.10 $\pm$ 1.73 & \textbf{99.93 $\pm$ 0.09} & 97.92 $\pm$ 1.72 & 98.71 $\pm$ 0.72 & 95.61 $\pm$ 2.88 \\
15 & 63.65 $\pm$ 6.29 & 82.41 $\pm$ 5.67 & 88.90 $\pm$ 1.10 & 72.18 $\pm$ 4.97 & 76.33 $\pm$ 4.62 & 87.23 $\pm$ 3.77 & 95.71 $\pm$ 4.57 & 81.86 $\pm$ 5.26 & \textbf{97.26 $\pm$ 1.63} \\
16 & 91.78 $\pm$ 3.19 & 96.89 $\pm$ 2.19 & 99.78 $\pm$ 0.19 & 98.45 $\pm$ 0.83 & 97.99 $\pm$ 0.61 & 98.47 $\pm$ 0.81 & 98.18 $\pm$ 2.92 & 98.99 $\pm$ 0.63 & \textbf{99.85 $\pm$ 0.06} \\ \hline
OA & 77.03 $\pm$ 1.26 & 80.14 $\pm$ 0.86 & 88.03 $\pm$ 1.26 & 90.35 $\pm$ 1.67 & 87.64 $\pm$ 0.83 & 92.82 $\pm$ 1.00 & 97.25 $\pm$ 0.87 & 92.67 $\pm$ 0.77 & \textbf{97.49 $\pm$ 1.14} \\
AA & 72.52 $\pm$ 1.41 & 86.99 $\pm$ 0.56 & 90.52 $\pm$ 1.41 & 95.72 $\pm$ 0.41 & 94.55 $\pm$ 0.57 & 96.12 $\pm$ 0.82 & \textbf{98.21 $\pm$ 0.30} & 95.69 $\pm$ 0.34 & 97.87 $\pm$ 0.85 \\
Kappa & 74.55 $\pm$ 2.07 & 76.52 $\pm$ 0.95 & 86.55 $\pm$ 2.07 & 89.26 $\pm$ 1.30 & 86.72 $\pm$ 1.01 & 93.26 $\pm$ 0.76 & 96.94 $\pm$ 0.96 & 92.06 $\pm$ 0.85 & \textbf{97.43 $\pm$ 0.79} \\ \hline
\end{tabular}}
\end{table*}

\subsection{Quantitative Evaluation}
Tables \ref{tb_jiangxia}-\ref{tb_salinas} present the test results of the Jiang Xia dataset, Xin Jiang dataset, and Salinas dataset, respectively, and bold is the best performance. The tables contain the prediction accuracy of each class and use OA, AA, and Kappa to evaluate the overall prediction standard.
We objectively analyzed the advantages and disadvantages of the algorithm using a quantitative analysis. The results showed that the methods based on the CNN and GCN have their own merits in terms of performance.
Overall, the 1D-CNN algorithm performs poorly because it is only based on spectral information. Methods such as CNN use the neighborhood blocks of image pixels as the input of the network and, to a certain extent, jointly utilize the spatial and spectral information of HSI.

By contrast, the overall classification accuracy of the $\text{S}^2$RC-GCN proposed in this study was the highest, indicating its effectiveness in classifying complex scenes. The OA, AA, and Kappa of the model on the Jiang Xia dataset all reached their highest values, which are 87.39\% $\pm$ 1.24\%, 86.69\% $\pm$ 0.88\% and 84.90\% $\pm$ 1.74\%, respectively; these values are 4.59\%, 3.89\%, and 5.21\% higher than ResCapsNet, respectively. The model demonstrated 71.42\% $\pm$ 0.56\%, 68.98\% $\pm$ 0.87\%, and 65.02\% $\pm$ 0.76\% accuracies, respectively in the Xin Jiang dataset. In terms of OA metrics, it outperforms the optimal models in CNN-based and GCN-based by 0.54\% and 2.62\%, respectively. This result shows that combining the feature information of the HSI with the spatial structure features can further improve classification performance. Despite the difficulty of testing on a complex land cover classification dataset, our proposed model nonetheless performed favorably compared to the other models in terms of classification performance. This demonstrates that our model has strong robustness. 

Table \ref{tb_salinas} illustrates the performance of each model on the generic hyperspectral dataset, showcasing that our model consistently achieves optimal results. Nevertheless, an interesting trend emerges where the CNN-based model outperforms the GCN-based model on the complex land cover classification dataset. In contrast, the phenomenon is reversed on the Salinas dataset. We speculate that this discrepancy arises due to the intricate nature of land cover datasets, necessitating models with enhanced feature mining capabilities. Given that our model integrates deep features and capitalizes on the strengths of the GCN model, the combination has been proven to be optimal for each dataset.

\begin{figure}
\label{pic:jiangxia}
\begin{center}
  \includegraphics[width=\linewidth]{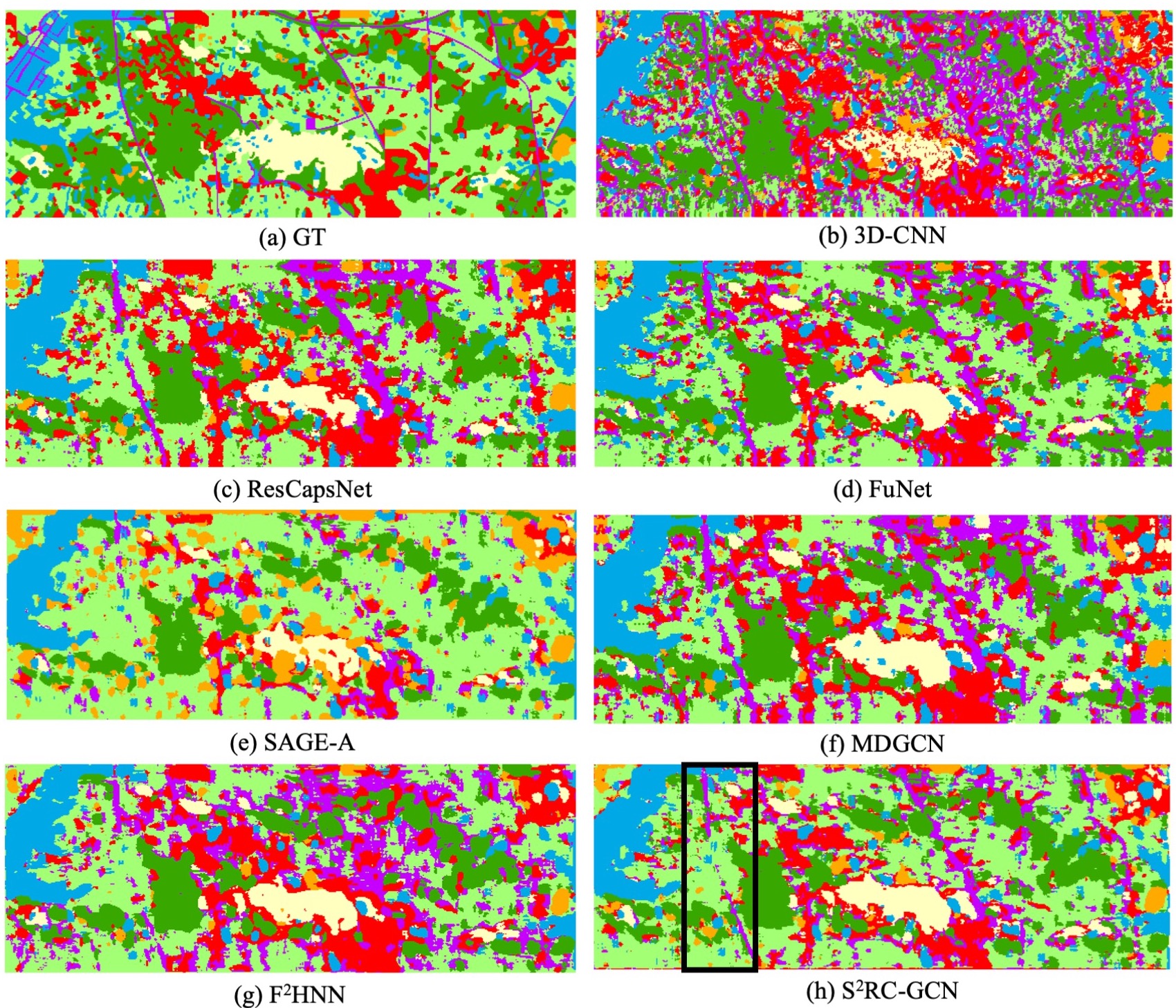}
\end{center}
\caption{The classification maps of different methods on the Jiang Xia dataset.}
\label{fig:jiangxia}
\end{figure}

\begin{figure*}
\label{pic:salinas}
\begin{center}
  \includegraphics[width=\linewidth]{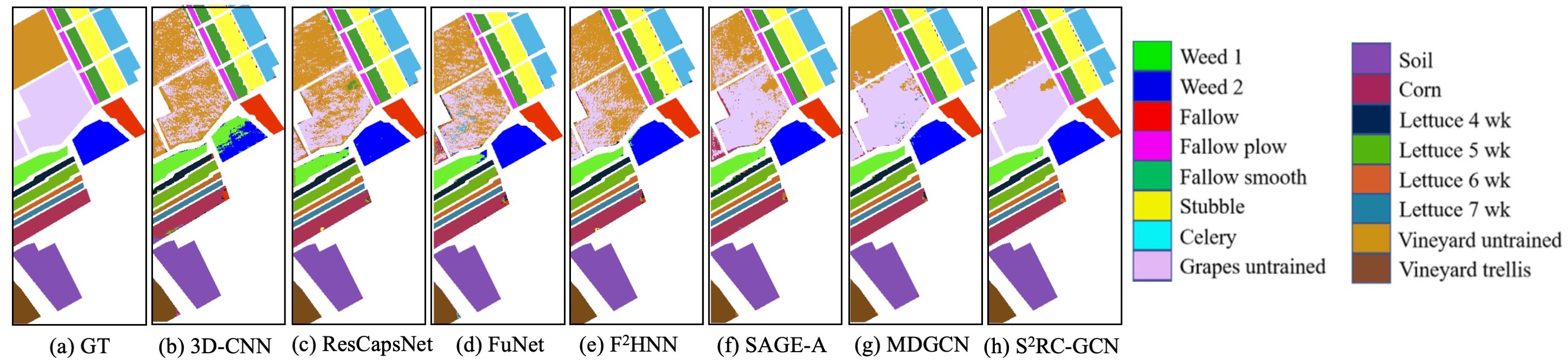}
\end{center}
\caption{The classification maps of different methods on the Salinas dataset.}
\label{fig:salinas}
\end{figure*}

\subsection{Classification Maps}
To observe the classification effect of the $\text{S}^2$RC-GCN method more intuitively, Figures \ref{fig:jiangxia} and \ref{fig:salinas} show the visual maps of different models for the Jiang Xia and Salinas, respectively. We can see that the maps of the entire study area obtained by the $\text{S}^2$RC-GCN method had superior classification accuracy and visual robustness. As indicated by the black squares in Figure \ref{fig:jiangxia}(h), the $\text{S}^2$RC-GCN model did not misclassify road areas as mineralized areas or bare land. $\text{S}^2$RC-GCN has a smoother classification, which more effectively divides the class boundaries of the mining area. In the overall classification of the Salinas, the $\text{S}^2$RC-GCN model exhibited a superior accuracy in the fine classification of grapes untrained and vineyard untrained.

In comparing the prediction results of the CNN-based and GCN-based algorithms for the entire image, these GCN-based methods have fewer misclassified points in the class boundary region. The remote sensing images of the two study areas have more complex ground-object distributions and spatial layout information. The fixed-structure CNN operator typically causes blurred edges, whereas the flexible-structure graph convolution operator is more suitable for pixel-level HSI classification.

\begin{figure}
\begin{center}
  \includegraphics[width=\linewidth]{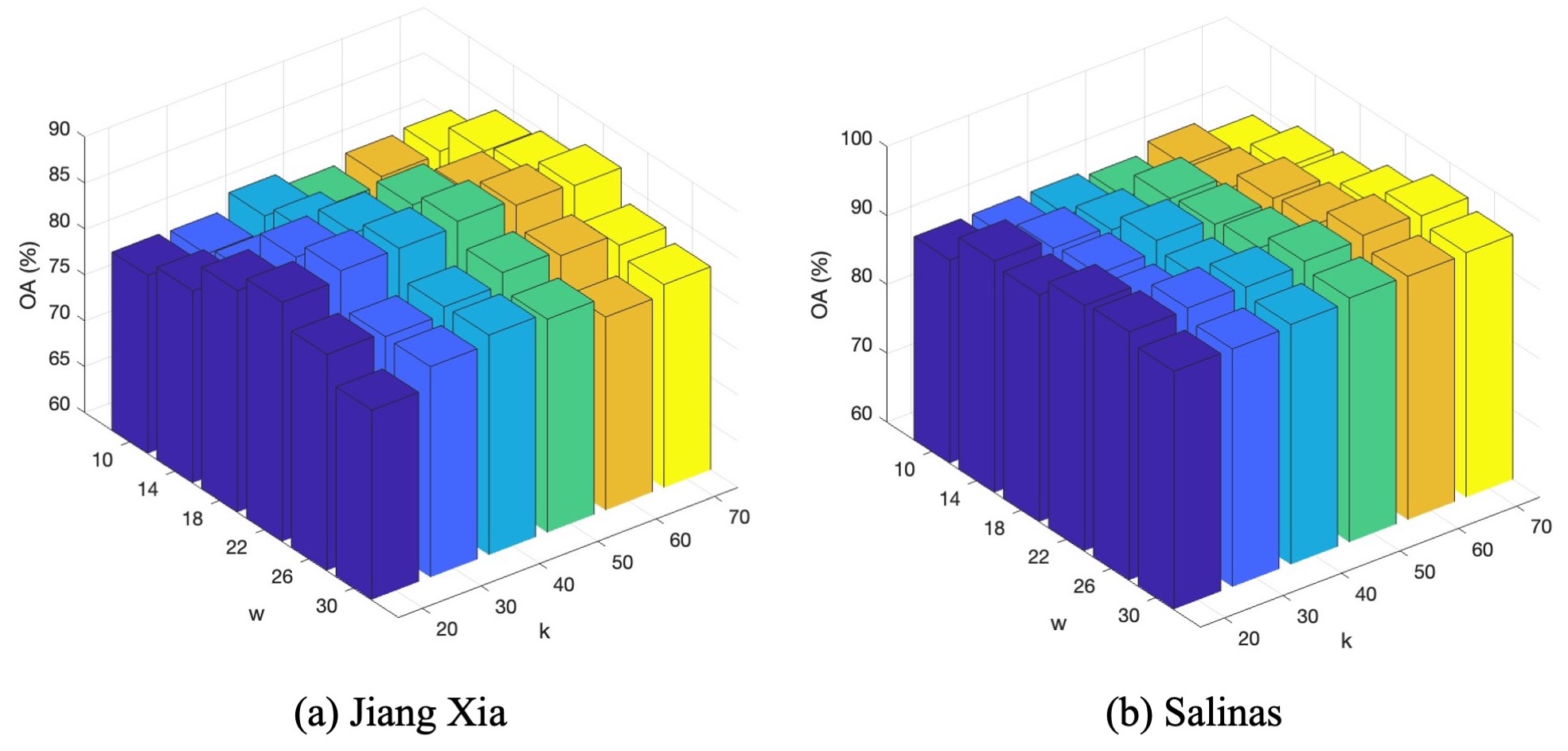}
\end{center}
\caption{Sensitivity to the $k$ and $w$ parameters on three datasets. $k$ and $w$, respectively, represent the value of k-nearest neighbor composition and the input size.}
\label{fig:para}
\end{figure}

\subsection{Parameter Analysis}

Figure \ref{fig:para} illustrates the impact of the number of neighbors $k$ and the input size $w$ on the OA across the Jiang Xia and Salinas datasets. Observing results, it becomes apparent that as $w$ increases, the overall accuracy initially shows improvement. This is attributed to the fact that a too small input size can hinder the capture of surrounding spatial information. However, the model does not consistently exhibit an upward trend. Upon reaching a critical value, the model accuracy begins to decline, suggesting an excess capture of redundant information. The number of nearest neighbors, denoted as $k$, signifies the count of neighbors in the adjacency matrix for each node. The proposed model employs aggregated feature composition to construct a more accurate adjacency matrix, aiming to reduce connections between nodes of different classes. As $k$ gradually increases, each node in the adjacency matrix incorporates more neighbors. However, these neighboring nodes may not belong to the same class as the focal node, and inputting this adjacency matrix into the GCN can impact classification outcomes. The figure demonstrates the model's resilience to changes in the value of $k$.

\begin{table}[]
\centering
\caption{Ablaton study, where (I) devotes a variant without fusion feature; (II) denotes replacing SE-block inside the 2D-CNN; (III) devotes a variant that removes the reliable contrastive graph convolution module.}
\label{tb_ablation}
\setlength{\tabcolsep}{3.5mm}{
\renewcommand\arraystretch{1.3}
\begin{tabular}{cccc} \hline
Methods & Jiang Xia & Xin Jiang & Salinas \\ \hline
(\uppercase\expandafter{\romannumeral1}) & 84.87 $\pm$ 1.28 & 69.44 $\pm$ 1.12 & 94.59 $\pm$ 1.28 \\
(\uppercase\expandafter{\romannumeral2}) & 86.93 $\pm$ 0.93 & 70.49 $\pm$ 0.62 & 96.93 $\pm$ 0.82 \\
(\uppercase\expandafter{\romannumeral3}) & 83.89 $\pm$ 0.81 & 70.23 $\pm$ 0.76 & 95.24 $\pm$ 0.94 \\
Ours & 87.39 $\pm$ 1.24 & 71.42 $\pm$ 0.56 & 97.49 $\pm$ 1.14 \\ \hline
\end{tabular}}
\end{table}

\subsection{Ablation Study}

In this section, we conduct extensive ablation experiments to verify the effectiveness of the proposed model. Specifically, we compare three variants of the $\text{S}^2$RC-GCN, (I) devotes a variant of our model that uses spectral and spatial features for contrastive learning rather than fusing features; (II) denotes replacing the attentional feature extraction module (SE-block) inside the 2D-CNN and using only regular ResNet to extract features; (III) devotes a variant of our model that removes the reliable contrastive graph convolution module. From the results in Table \ref{tb_ablation}, we can draw the following conclusions:

(1) The joint feature obtained by integrating spectral and spatial features exhibit superior performance compared to the spatial features alone. This superiority arises from the higher feature similarity between the joint feature and the spatial feature, aligning with the enhancement requirements in the comparison model and ensuring a strong connection between the two.

(2) The lack of the SE-block in the 2D-CNN will lead to a decrease in accuracy because the attention-based feature selection module can select more favorable features and filter the noisy information.

(2) By comparing our model with the results of (\uppercase\expandafter{\romannumeral3}), we can observe that the reliable contrastive graph convolution we proposed can reduce redundant information and improve the accuracy of the model. The omission of the attention mechanism led to reduced classification results, with an OA drop of 3.50\%, 1.19\%, and 1.85\% for the three datasets, respectively.

\section{Conclusion}

To improve the classification accuracy of HSI in complex backgrounds, this paper proposes a spatial-spectral reliable contrastive graph convolutional network named $\text{S}^2$RC-GCN. $\text{S}^2$RC-GCN uses 1D-CNN and 2D-CNN combined with SE attention module to extract important information to obtain spatial-spectral features. To make the joint feature more discriminative, this paper designs an adaptive group graph convolution module to improve the feature representation of the graph. In addition, we propose a reliable contrastive GCN, which uses nodes with true labels or predictive labels with high thresholds for contrastive learning to further learn more efficient graph representations. The experimental results show that $\text{S}^2$RC-GCN can effectively improve the classification performance of complex remote sensing images. In future research, we will consider the multimodal features of HSI to further improve the classification accuracy of fine land cover.

\section{Acknowledgment}
This study was jointly supported by the Natural Science Foundation of China under Grants 42071430 and U21A2013, the Opening Fund of Key Laboratory of Geological Survey and Evaluation of Ministry of Education under Grant GLAB2022ZR02 and Grant GLAB2020ZR14.

Computation of this study was performed by the High-performance GPU Server (TX321203) Computing Centre of the National Education Field Equipment Renewal and Renovation Loan Financial Subsidy Project of China University of Geosciences, Wuhan.

%
%


\bibliographystyle{IEEEtran}
\bibliography{IEEEfull,refs}

%
%



\end{document}